\documentclass[11pt,a4paper]{article}
\usepackage[table]{xcolor}
\usepackage[hyperref]{acl2019}
\usepackage{times}
\usepackage{latexsym}
\usepackage{amsmath}
\usepackage{amsfonts}
\usepackage{url}
\usepackage{comment}
\usepackage{graphicx}

\newcommand{\grow}[1]{\rowcolor{gray!10} #1}

\aclfinalcopy % Uncomment this line for the final submission
\DeclareMathOperator*{\argmin}{argmin}

\newcommand{\minisection}[1]{\noindent{\bf #1}\nobreak}

\expandafter\def\expandafter\normalsize\expandafter{%
    \normalsize
    \setlength\abovedisplayskip{2pt} % reduce space around math
    \setlength\belowdisplayskip{2pt}
    \setlength\abovedisplayshortskip{2pt}
    \setlength\belowdisplayshortskip{2pt}
    \setlength{\textfloatsep}{0.6\textfloatsep} % reduce space around figures
    \setlength{\floatsep}{0.6\floatsep}
    \setlength{\intextsep}{0.6\intextsep}
}

\title{Unsupervised Paraphrasing without Translation}

\author{Aurko Roy \\
  Google Research \\
  \texttt{aurkor@google.com} \\\And
  David Grangier \\
  Google Research \\
  \texttt{grangier@google.com} \\}

\date{}

\begin{document}
\maketitle
\begin{abstract}
Paraphrasing exemplifies the ability to 
abstract semantic content from surface forms. Recent work 
on automatic paraphrasing is dominated by methods leveraging 
Machine Translation (MT) as an intermediate step. This contrasts with 
humans, who can paraphrase without being bilingual.
This work proposes to learn paraphrasing models from an unlabeled 
monolingual corpus only. To that end, we propose a residual variant of 
vector-quantized variational auto-encoder.

We compare with MT-based approaches on paraphrase identification, 
generation, and training augmentation. 
Monolingual paraphrasing outperforms unsupervised translation
in all settings. Comparisons with supervised translation are more mixed: 
monolingual paraphrasing is interesting for identification and 
augmentation; supervised translation is superior for generation.
\end{abstract}

\section{Introduction}

Many methods have been developed to generate paraphrases automatically~\cite{madnani2010}.
Approaches relying on Machine Translation (MT)
have proven popular due to the scarcity of labeled paraphrase pairs~\cite{callison2007paraphrasing,mallinson2017paraphrasing,iyyer2018adversarial}.
Recent progress in MT with neural methods~\cite{bahdanau2014neural,transformer} has popularized this latter strategy. 
Conceptually, translation is appealing since it abstracts semantic content from its linguistic realization. 
For instance, assigning the same source sentence to multiple translators will result in a rich set of semantically close 
sentences~\cite{callison2007paraphrasing}. At the same time, bilingualism does not seem necessary to humans to generate paraphrases.

This work evaluates if data in two languages is necessary for 
paraphrasing. We consider three settings: supervised translation (parallel bilingual data is used), unsupervised 
translation (non-parallel corpora in two languages are used) and monolingual (only unlabeled data in the paraphrasing language is used). Our comparison devises comparable encoder-decoder neural networks for all three settings. While the literature on supervised~\cite{bahdanau2014neural,cho2014properties,transformer} and 
unsupervised translation~\cite{lample2017unsupervised,artetxe2018iclr,lample2018phrase} offer 
solutions for the bilingual settings, monolingual neural paraphrase generation
has not received the same attention.

We consider discrete and continuous auto-encoders in an unlabeled monolingual setting,
and contribute improvements in that context. We introduce a model based on 
Vector-Quantized Auto-Encoders, VQ-VAE~\cite{vqvae}, for generating 
paraphrases in a purely monolingual setting. Our model introduces residual connections parallel to the quantized 
bottleneck. This lets us interpolate from classical continuous auto-encoder~\cite{stackeddenoising} to VQ-VAE.
Compared to VQ-VAE, our architecture offers a better control over the decoder 
entropy and eases optimization. Compared to continuous auto-encoder, our 
method permits the generation of diverse, but semantically close 
sentences from an input sentence.

We compare paraphrasing models over intrinsic and extrinsic metrics. 
Our intrinsic evaluation evaluates paraphrase identification, 
and generations. Our extrinsic 
evaluation reports the impact of training augmentation with paraphrases on text 
classification. Overall, monolingual approaches can
outperform unsupervised translation in all settings. Comparison with supervised translation shows
that parallel data provides valuable information for paraphrase 
generation compared to purely monolingual training.

\section{Related Work}

\minisection{Paraphrase Generation}
Paraphrases express the same content with alternative surface forms. Their automatic generation has been studied for decades: rule-based~\cite{mckeown1980paraphrasing,meteer1988strategies} and data-driven methods~\cite{madnani2010} have been explored. Data-driven approaches have considered different source of training data, including multiple translations of the same text~\cite{barzilay2001extracting,pang2003} or alignments of comparable corpora, such as news from the same period~\cite{dolan2004,barzilay2003alignment}.

Machine translation later emerged as a dominant method for paraphrase generation. \citet{bannard2005paraphrasing} identify equivalent English phrases mapping to the same non-English phrases from an MT phrase table. \citet{kok2010} performs random walks across multiple phrase tables. Translation-based paraphrasing has recently benefited from neural networks for MT~\cite{bahdanau2014neural,transformer}. Neural MT can generate paraphrase pairs by translating one side of a parallel corpus~\cite{wieting2018paranmt,iyyer2018adversarial}. Paraphrase generation with pivot/round-trip neural translation has also been used~\cite{mallinson2017paraphrasing,yu2018qanet}.

Although less common, monolingual neural sequence models have also been proposed. In 
supervised settings, \citet{prakash2016neural, gupta2018deep} learn sequence-to-sequence models on paraphrase data. In 
unsupervised settings, \citet{textvae} apply a VAE to paraphrase detection while \citet{li2017paraphrase} train a 
paraphrase generator with adversarial training.

\minisection{Paraphrase Evaluation}
Evaluation can be performed by human raters, evaluating both text fluency and semantic similarity. Automatic evaluation 
is more challenging but necessary for system development and larger scale statistical analysis~\cite{callison2007paraphrasing,madnani2010}. 
Automatic evaluation and generation are actually linked: if an automated metric would reliably assess the semantic similarity and fluency of a pair of sentences, one would generate by searching the space of sentences to maximize that metric.
Automated evaluation can report the overlap with a reference paraphrase, like for translation \cite{papineni2002bleu} or summarization~\cite{lin2004rouge}. BLEU, METEOR and TER metrics have been used \cite{prakash2016neural, gupta2018deep}. 
These metrics do not evaluate whether the generated paraphrase differs from the input sentence and large amount of input 
copying is not penalized.
\citet{galley2015deltableu} compare overlap with multiple references, weighted by quality; 
while~\citet{sun2012joint} explicitly penalize overlap with the input sentence. 
\citet{grangier2017quickedit} alternatively compare systems which have first been calibrated to 
a reference level of overlap with the input. We follow this strategy and calibrate the generation 
overlap to match the average overlap observed in paraphrases from humans.

In addition to generation, probabilistic models can be assessed through scoring. 
For a sentence pair \((x, y)\), the model estimate of $P(y|x)$ can be used to discriminate between paraphrase and non-paraphrase 
pairs~\cite{dolan2005automatically}. The correlation of model scores with human judgments~\cite{cer2017semeval} can also be assessed. 
We report both types of evaluation.

Finally, paraphrasing can also impact downstream tasks, e.g. to generate additional training data by paraphrasing training 
sentences~\cite{marton2009improved,zhang2015character,yu2018qanet}. We evaluate this impact for classification tasks.

\section{Residual VQ-VAE for Unsupervised Monolingual Paraphrasing}
\label{sec:res-vq-vae}

Auto-encoders can be applied to monolingual paraphrasing. Our work 
combines Transformer networks~\cite{transformer} and 
VQ-VAE~\cite{vqvae}, building upon recent work in 
discrete latent models for translation~\cite{kaiser2018fast, 
roy2018theory}. VQ-VAEs, as opposed to 
continuous VAEs, rely on discrete latent variables. This is interesting 
for paraphrasing as it 
equips the model with an 
explicit control over the latent code capacity, allowing the model to 
group multiple related examples under the same 
latent assignment, similarly to classical clustering 
algorithms~\cite{kmeans}. 
This is conceptually simpler and more effective than rate 
regularization~\cite{higgins2016beta} or denoising 
objectives~\cite{stackeddenoising} for continuous auto-encoders.
At the same time, training auto-encoder with discrete bottleneck is 
difficult~\cite{roy2018theory}. 
We address this difficulty with an hybrid model using a continuous 
residual connection around 
the quantization module.

\begin{figure}[h]
\begin{center}
\includegraphics[width=0.40\textwidth]{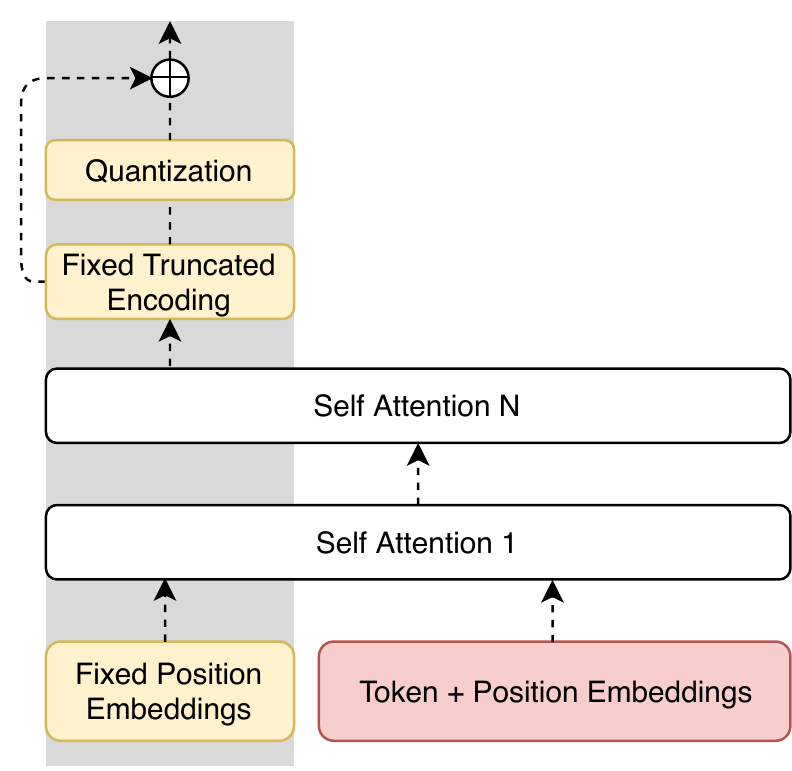} 
\end{center}
\caption{Encoder Architecture} \label{fig:encoder}
\end{figure}

We modify the Transformer encoder~\cite{transformer} as depicted in 
Figure~\ref{fig:encoder}.
Our encoder maps a sentence into a fixed size vector. This is simple 
and avoids choosing a fixed
length compression rate between the input and the latent 
representation~\cite{kaiser2018fast}.
Our strategy to produce a fixed sized representation from transformer
is analogous to the special token employed for sentence classification in~\citep{devlin2018bert}.

At the first layer, we extend
the input sequences with one or more fixed positions which are part of 
the self-attention stack. 
At the output layer, the encoder output is restricted to these special 
positions which constitute the encoder fixed sized-output. As in 
\cite{kaiser2018fast}, this vector is split into multiple heads 
(sub-vectors of equal dimensions) 
which each goes through a quantization module. For each head $h$, the 
encoder output $e_h$ 
is quantized as,
$$
q_h(e_h) = c_k, \textrm{~where~} k = \argmin_i \| e_h - c_i \|^2
$$
where $\{c_i\}_{i=0}^K$ denotes the codebook vectors. 
The codebook is shared across heads and training combines straight-through gradient estimation and exponentiated 
moving averages~\cite{vqvae}. The quantization module is completed with a residual connection, 
with a learnable weight $\alpha$,
$
z_h(e_h) = \alpha e_h + (1 - \alpha) q_h(e_h).
$
One can observe that residual vectors and quantized vectors always have similar norms by 
definition of the VQ module. This is a fundamental difference with classical continuous 
residual networks, where the network can reduce activation norms of some modules to 
effectively rely mostly on the residual path.
This makes $\alpha$ an important parameter to trade-off continuous and discrete auto-encoding.
Our learning encourages the quantized path with a squared penalty $\alpha^2$.

After residual addition, the multiple heads of the resulting vector are 
presented as a matrix to which 
a regular transformer decoder can attend. Models are trained to maximize the 
likelihood of the training set 
with Adam optimizer using the learning schedule from~\cite{transformer}.

\begin{table*}[ht!]
    \centering
    %\resizebox{\columnwidth}{!}{
    \begin{tabular}{|l | c | c | c | c| c|}
    \hline
       & \multicolumn{3}{c|}{Parapharase Identification} & \multicolumn{2}{c|}{Generation}\\
       & MRPC & STS     & MTC     & BLEU & Pref. \\\hline
        Supervised Translation   & 70.6 & 46.0 & 78.6 & \textbf{8.73} & 36.8 \\ 
        + Distillation    & 66.5 & \textbf{60.0} & 55.6 & 7.08 & -- \\\hline
        Unsupervised Translation & 66.0 & 13.2 & 65.8 & 6.59 & 28.1 \\ 
        + Distillation & 66.9 & 45.0 & 52.0 & 6.45 & --\\\hline
        Mono. DN-AE & 66.8 & 46.2 & 91.6 & 5.13 & -- \\ \hline
        Mono. VQVAE& 66.3 & 10.6 & 69.0 & 3.85 & -- \\ 
        + Residual      & \textbf{73.3}& 59.8 & \textbf{94.0} & 7.26 & 31.9 \\ 
        + Distillation  & 71.3 & 54.3 & 88.4 & 6.88 & -- \\ \hline     
    \end{tabular}
    %}
    \caption{Paraphrase Identification \& Generation. Identification is evaluated with 
    accuracy on MRPC, Pearson Correlation on STS and ranking on MTC. Generation is evaluated
    with BLEU and human preferences on MTC.}
    \label{tab:identification}
\end{table*}

\begin{table*}[!htb]
    \centering
    %\resizebox{\columnwidth}{!}{
    \begin{tabular}{|l | c c | c c |}
    \hline
       &\multicolumn{2}{c|}{SST-2} & \multicolumn{2}{c|}{TREC} \\
                         & Acc. & F1 & Acc & F1 \\\hline
        NB-SVM (trigram) &  81.93  & 83.15  &     89.77 & 84.81 \\\hline
        Supervised Translation    &  81.55  & 82.75  & {\bf 90.78} & 85.44 \\ 
        + Distillation   & 81.16 & 66.59 & 90.38 & {\bf 86.05} \\ \hline
        Unsupervised Translation  & 81.87 & 83.18 & 88.17 & 83.42\\ 
        + Distillation   & 81.49 & 82.78 & 89.18 & 84.41 \\ \hline
        Mono. DN-AE & 81.11 & 82.48 & 89.37 &  84.08 \\
        Mono. VQ-VAE & 81.98 & 82.95 & 89.17 & 83.64 \\ 
        + Residual       & {\bf 82.12}& {\bf 83.23} & 89.98 & 84.31 \\ 
        + Distillation   & 81.60 & 82.81 & 89.78 & 84.31 \\ 
        \hline
    \end{tabular}
    %}
    \caption{Paraphrasing for Data Augmentation: Accuracy and F1-scores of a Naive Bayes-SVM 
    classifier on sentiment (SST-2) and question (TREC) classification.}
    \label{tab:augmentation}
\end{table*}

\section{Experiments \& Results}

We compare neural paraphrasing with and without access to bilingual data. For bilingual settings, 
we consider supervised and unsupervised translation using round-trip translation~\cite{mallinson2017paraphrasing,yu2018qanet} 
with German as the pivot language. 
Supervised translation trains the transformer base model~\cite{transformer} on 
the WMT'17 English-German parallel data~\cite{bojar-EtAl:2017:WMT1}. Unsupervised translation
considers a pair of comparable corpora for training, German and English WMT-Newscrawl 
corpora, and relies on the transformer models from~\citet{lample2018phrase}.
Both MT cases train a model from English to German and from German to 
English to perform round-trip MT. For each model, we also distill the round-trip 
model into a single artificial English to English model by generating a training 
set from pivoted data. Distillation relies on the billion word corpus, 
LM1B~\cite{lm1b}. 

Monolingual Residual VQ-VAE is trained only on LM1B with \(K = 2^{16}\), with \(2\) 
heads and fixed window of size \(16\). We also evaluate plain VQ-VAE
\(\alpha = 0\) to highlight the value of our residual modification.
We further compare with a monolingual continuous denoising auto-encoder (DN-AE), 
with noising from~\citet{lample2018phrase}.

\minisection{Paraphrase Identification}
For classification of sentence pairs $(x,y)$ over 
Microsoft Research Paraphrase Corpus (MRPC) from \citet{dolan2005automatically}, we 
train logistic regression on $P(y|x)$ and $P(x|y)$ from the model, 
complemented with encoder outputs in fixed context settings. We also 
perform paraphrase quality regression on Semantic Textual Similarity (STS) from \citet{cer2017semeval} by 
training ridge regression on the same features.

Finally, we perform paraphrase ranking on Multiple Translation Chinese (MTC) from \citet{huang2002multiple}. MTC contains English 
paraphrases collected as translations of the same Chinese sentences from multiple 
translators~\cite{mallinson2017paraphrasing}. We pair each MTC sentence $x$ with a paraphrase $y$ 
and 100 randomly chosen non-paraphrases $y'$. We compare the paraphrase score $P(y|x)$ to 
the $100$ non-paraphrase scores $P(y'|x)$ and report the fraction of comparisons where the 
paraphrase score is higher.

Table~\ref{tab:identification} (left) reports that our residual model 
outperforms alternatives in all identification setting, except for STS,
where our Pearson correlation is slightly under supervised translation.

\minisection{Paraphrases for Data Augmentation}
We augment the training set of text classification tasks for sentiment analysis on Stanford Sentiment Treebank (SST-2)~\citep{socher2013recursive} 
and question classification on Text REtrieval Conference 
(TREC)~\citep{voorhees2000building}. In both cases, we double training set size by
paraphrasing each sentence and train Support Vector Machines with Naive Bayes features~\cite{wang2012baselines}.

In Table~\ref{tab:augmentation}, augmentation with monolingual models yield the best performance
for SST-2 sentiment classification. TREC question classification is better with supervised translation augmentation. 
Unfortunately, our monolingual training set LM1B does not contain many question sentences. Future work
will revisit monolingual training on larger, more diverse resources.

\begin{table*}[ht]
    \centering
    %\resizebox{\columnwidth}{!}{
    \begin{tabular}{|l|p{1.95\columnwidth}|}
    \hline
    \grow \textbf{\small{In:}} & a worthy substitute \\
    \textbf{\small{Out:}} & A worthy replacement.\\\hline
    \grow \small{\textbf{In:}} & Local governments will manage the smaller enterprises.\\
    \textbf{\small{Out:}} & Local governments will manage smaller companies.\\\hline
    \grow \textbf{\small{In:}} & Inchon is 40 kilometers away from the border of North Korea.\\
    \textbf{\small{Out:}} & Inchon is 40 km away from the North Korean border.\\\hline
    \grow \textbf{\small{In:}} & Executive Chairman of Palestinian 
    Liberation Organization, Yasar Arafat, and other leaders are often critical 
    of aiding countries not fulfilling their promise to provide funds in a 
    timely fashion.\\
    \textbf{\small{Out:}} & Yasar Arafat , executive chairman of the Palestinian Liberation Organization and other leaders are often critical of helping countries meet their pledge not to provide funds in a timely fashion.\\
    \hline
       \end{tabular}
     %  }
    \caption{Examples of generated paraphrases from the monolingual residual model (Greedy search).}
    \label{tab:examples}
\end{table*}

\minisection{Paraphrase Generation}
Paraphrase generation are evaluated on MTC. We select the 4 best 
translators according to MTC documentation and paraphrase pairs 
with a length ratio under $1.2$. 
Our evaluation prevents trivial copying solutions. We select 
sampling temperature for all models such that their generation 
overlap with the input is 20.9 BLEU, the average overlap between 
humans on MTC. We report BLEU overlap with the target and run a blind 
human evaluation where raters pick the best generation
among supervised translation, unsupervised translation and monolingual.

Table~\ref{tab:examples} shows examples.
Table~\ref{tab:identification} (right) reports that monolingual paraphrasing compares favorably
with unsupervised translation while supervised translation is the best technique. This highlights the value
of parallel data for paraphrase generation. 

\section{Discussions}
Our experiments highlight the importance of the residual
connection for paraphrase identification. From 
Table~\ref{tab:identification}, we see that a model without the
residual connection obtains \(66.3\%, 10.6\%\) and 
\(69.0\%\) accuracy on MRPC, STS and MTC. Adding 
the residual connection improves this to \(73.3\%, 59.8\%\)
and \(94.0\%\) respectively.

The examples in Table~\ref{tab:examples} show 
paraphrases generated by the model. The overlap with the input
from these examples is high. It is possible to generate sentences
with less overlap at higher sampling temperatures, we however observe
that this strategy impairs fluency and adequacy. We plan to explore 
strategies which allow to condition the decoding process on 
an overlap requirement instead of varying sampling temperatures
\citep{grangier2017quickedit}.

\section{Conclusion}

We compared neural paraphrasing with and without access to bilingual data. Bilingual 
settings considered supervised and unsupervised translation. Monolingual settings considered
auto-encoders trained on unlabeled text and introduced continuous residual connections 
for discrete auto-encoders. This method is advantageous over both discrete and continuous 
auto-encoders. Overall, we showed that monolingual models can outperform bilingual 
ones for paraphrase identification and data-augmentation through paraphrasing. We also 
reported that generation quality from monolingual models can be higher than model based on
unsupervised translation but not supervised translation. Access to parallel data is therefore still advantageous
for paraphrase generation and our monolingual method can be a helpful resource for
languages where such data is not available.

\section*{Acknowledgments}

We thanks the anonymous reviewers for their suggestions. We thank the
authors of the Tensor2tensor library used in our experiments~\cite{vaswani2018tensor2tensor}.
%The acknowledgments should go immediately before the references.  Do
%not number the acknowledgments section. Do not include this section
%when submitting your paper for review. \\

\bibliography{deeplearn}
\bibliographystyle{acl_natbib}

\end{document}